\documentclass[11pt]{article}

\usepackage{acl}

\usepackage{times}
\usepackage{latexsym}
\usepackage{enumitem}
\usepackage{booktabs} 

\usepackage[T1]{fontenc}

\usepackage[utf8]{inputenc}

\usepackage{microtype}

\usepackage{inconsolata}

\usepackage{graphicx}

\title{Bridging Human and Model Perspectives: A Comparative Analysis of Political Bias Detection in News Media Using Large Language Models}

\author{
Shreya Adrita Banik, Niaz Nafi Rahman, Tahsina Moiukh, Farig  Sadeque \\
  Department of Computer Science and Engineering, BRAC University \\
  \texttt{\small \{shreya.adrita.banik, niaz.nafi.rahman, tahsina.moiukh\}@g.bracu.ac.bd  farig.sadeque@bracu.ac.bd}
}

\begin{document}
\maketitle
\begin{abstract}
Detecting political bias in news media is a complex task that requires interpreting subtle linguistic and contextual cues. Although recent advances in Natural Language Processing (NLP) have enabled automatic bias classification, the extent to which large language models (LLMs) align with human judgment still remains relatively underexplored and not yet well understood. This study aims to present a comparative framework for evaluating the detection of political bias across human annotations and multiple LLMs, including GPT, BERT, RoBERTa, and FLAN. We construct a manually annotated dataset of news articles and assess annotation consistency, bias polarity, and inter-model agreement to quantify divergence between human and model perceptions of bias. Experimental results show that among traditional transformer-based models, RoBERTa achieves the highest alignment with human labels, whereas generative models such as GPT demonstrate the strongest overall agreement with human annotations in a zero-shot setting. Among all transformer-based baselines, our fine-tuned RoBERTa model acquired the highest accuracy and the strongest alignment with human-annotated labels. Our findings highlight systematic differences in how humans and LLMs perceive political slant, underscoring the need for hybrid evaluation frameworks that combine human interpretability with model scalability in automated media bias detection.
\end{abstract}

\section{Introduction}

Political bias in news discourse shapes public opinion and influences democratic decision-making. 
While recent advances in transformer-based language models have improved automated bias detection, 
the extent to which these models \textit{interpret} bias in ways that align with human reasoning remains unclear. 
Most prior work focuses on training classifiers or analyzing ideological framing, treating human annotation as a fixed ground truth. 
However, this overlooks a key question: \textit{do models and humans rely on similar cues when identifying political bias?}

In this work, we introduce a comparative framework to evaluate human–model alignment in political bias annotation. 
We analyze how different model families like  BERT, RoBERTa, XLM-R, FLAN-T5, and GPT-5 assign bias labels under zero-shot, guideline-conditioned, and few-shot prompting regimes. 
Using a human-annotated corpus of news text, we examine not only model accuracy, but also the interpretive strategies underlying model decisions. 
Our analysis shows clear differences, like relying too much on word sentiment cues and struggling to tell apart quoted speech from authorial stance.

The results indicate that while fine-tuned transformers perform well on explicit bias signals, generative models like GPT-5 show better zero-shot alignment with human annotation patterns. 
Yet across all model families, alignment weakens for implicit or context-dependent cases. This suggests that current systems detect \textit{entity-level sentiment} rather than the \textit{discourse-level framing} used in human judgment. 
This gap highlights the need for bias detection methods that include contextual inference, speaker attribution, and narrative structure.

Our contributions are threefold:
\begin{enumerate}
    \item \textbf{Human-Annotated Bias Corpus.} We construct a human-annotated corpus of news articles labeled for political bias which involves a multi-step annotation process with adjudicated disagreement resolution. This approach offers an interpretable and consistently coded benchmark for examining human--model alignment in bias perception.

    \item \textbf{Unified Evaluation Framework.} We present a unified framework that compares annotation behavior across multiple model families (BERT, RoBERTa, XLM-R, FLAN-T5, and GPT-based models) under zero-shot, guideline-conditioned, and few-shot prompting configurations. The framework separates the effects of model architecture from those of inference strategy on bias classification.

    \item \textbf{Qualitative Divergence Analysis.} We provide a comparative analysis that shows systematic divergences between human reasoning and model predictions, including:
    (i) reliance on lexical polarity rather than contextual framing,
    (ii) failure to differentiate between speaker attribution from author stance,
    and (iii) entity-centered heuristics that overshadow multi-actor narrative interpretation.
\end{enumerate}

\noindent Taken together, our findings show that higher model accuracy does not imply human-aligned interpretation. Current models perform bias classification primarily through sentiment- and entity-level heuristics, whereas human annotators rely on discourse structure, contextual justification, and pragmatic inference. This shows a significant difference between pattern-matching behavior and interpretive reasoning in detecting political bias.

\section{Related Work}

\subsection{Political Bias Detection in NLP}
Political bias detection has been a central topic in computational social science and NLP, aiming to identify ideological slant and framing in textual media. Moverover, modern Language Models have political learnings that reinforce polarization (\cite{feng2023pretrainingdatalanguagemodels}).
Early research employed linguistic and source-level features to classify bias in news reporting \citep{baly2020detect, fan2019hierarchical, morstatter2018bias}. However media bias is a multi-dimensional phenomenon manifested through framing, agenda-setting, and information selection, yet it is often operationalized in computational research as a unidimensional text-classification task (\cite{rodrigo2024media}).
Recent transformer-based models have improved robustness by leveraging contextualized embeddings for political ideology prediction. 
However, these approaches rely heavily on human-labeled data, and the subjectivity inherent in political labeling limits the generalizability and interpretability of the resulting models.

\subsection{Bias and Fairness in Large Language Models}
Large Language Models (LLMs) have transformed NLP evaluation, revealing both their potential as detectors of social bias and their own representational biases.
\citet{jaremko2025implicit} show that GPT-4 and LLaMA-3 achieve near-human performance in identifying implicitly abusive language. This indicates that LLMs capture nuanced linguistic features correlated with implicit bias. 
Similarly, \citet{movva2024alignment} evaluate annotation alignment between GPT-4 and human annotators across conversational safety datasets and their findings include that GPT-4 often matches or exceeds the median human–human correlation. 
\citet{muenker2025zeroshot} demonstrate that zero-shot prompt-based annotation using models such as FLAN-T5 or mT0 can achieve results comparable to fine-tuned classifiers on multilingual political datasets. 
These findings collectively suggest that LLMs can serve as capable, if imperfect, substitutes for human annotators, particularly for complex or subjective linguistic phenomena.

\subsection{Human--Machine Annotation Alignment}

Recent studies have increasingly examined how closely large language model (LLM) annotations align with human judgment, particularly for subjective NLP tasks. Prior work highlights substantial variability among human annotators themselves, as well as between humans and models \citep{plank2022human, davani2022disagreement}. \citet{chiang2023evaluation} and \citet{gilardi2023gpt} find that GPT-based models can replicate or even outperform crowdworkers in text evaluation tasks, with GPT-3.5 producing higher-quality annotations than MTurk at significantly lower cost. Similarly, \citet{movva2024alignment} report that GPT-4 achieves higher correlation with average human ratings than the median human annotator. \cite{bojic2025} find GPT-4 yields greater inter-rater reliability than humans on political stance and sentiment annotation. Prompt-based models such as FLAN-T5 and mT0 have also demonstrated competitive zero-shot annotation performance in political discourse and topic classification \citep{jaremko2025implicit, muenker2025zeroshot}.

Despite these advances, alignment remains imperfect. \citet{santurkar2023opinions} show that LLM outputs often reflect a homogenized ideological bias (e.g., left-leaning), while \citet{lee2023opiniongaps} observe that current models fail to fully capture the diversity of human opinions. In the fairness domain, multilingual models trained on stereotype benchmarks like CrowS-Pairs and BBQ show reduced bias and improved generalization compared to monolingual counterparts \citep{nie2024multilingual}. These findings collectively motivate our study, which systematically compares human and model annotations in political bias detection, quantifying divergence across multiple model families, including GPT, BERT, RoBERTa, XLM-R, and FLAN.

\section{Dataset} \label{sec:dataset}

\subsection{Dataset Overview}
We use the \textbf{FIGNEWS 2024} dataset, released as part of the ArgMining shared task on news media narratives \citep{fignews2024,zaghouani2024fignews}. The dataset consists of news headlines and Facebook posts related to the Israel--Palestine conflict, published between \textit{October 7, 2023} and \textit{January 31, 2024}. It includes data in five languages---English, Arabic, French, Hebrew, and Hindi---with machine translations to English provided by the organizers. For consistency and comparability, all entries were analyzed in English.

After removing irrelevant metadata and incomplete entries, our final dataset contains approximately \textbf{15,000 samples} with 11 attributes, later reduced to 5 core columns (\texttt{source}, \texttt{type}, \texttt{text}, \texttt{translation}, and \texttt{bias}). Each instance was manually annotated for bias into one of seven labels: \textit{Unclear}, \textit{Biased Against Palestine}, \textit{Biased Against Israel}, \textit{Unbiased}, \textit{Biased Against Both}, \textit{Biased Against Others}, and \textit{Not Applicable}. The data was divided into \textbf{70\% training}, \textbf{10\% validation}, and \textbf{20\% test} splits.

\subsection{Human Annotation Process}
The annotation was performed manually by three researchers following a structured annotation guideline inspired by the FIGNEWS shared task objectives. Each annotator labeled entries independently, after which discrepancies were resolved through group discussion. Inter-annotator agreement (IAA) was calculated using Cohen’s $\kappa$, yielding substantial agreement scores of \textbf{0.65--0.68}. 

Annotators assessed whether each text was relevant to the conflict, objective or subjective, and if biased, toward which entity the bias was directed. Weekly consensus meetings ensured that ambiguous or borderline cases were discussed and resolved. The resulting manually labeled set serves as the \textit{human benchmark} for the comparative analysis presented in later sections.

\subsection{Annotation Guidelines}
The annotation schema defines seven mutually exclusive bias categories:

\begin{itemize}[itemsep=0pt, parsep=0pt, topsep=2pt, partopsep=0pt]
    \item \textbf{Unbiased}: Objective reporting that represents all sides fairly and avoids loaded language.
    \item \textbf{Unclear}: Texts lacking sufficient context to determine direction of bias.
        \item \textbf{Biased Against Israel}: Negative framing or delegitimization of Israel or its institutions.
    \item \textbf{Biased Against Palestine}: Negative framing or stereotypes directed at Palestinians or Palestinian groups.
    \item \textbf{Biased Against Both}: Simultaneous negative portrayal of both sides.
    \item \textbf{Biased Against Others}: Bias directed toward external actors (e.g., USA, UN, Iran).
    \item \textbf{Not Applicable}: Texts unrelated to the Israel--Palestine conflict.
\end{itemize}

Full annotation guidelines with examples for each class are provided in the Appendix ~\ref{sec:appendix}.

\begin{figure}[t]
\centering
\includegraphics[width=0.48\textwidth]{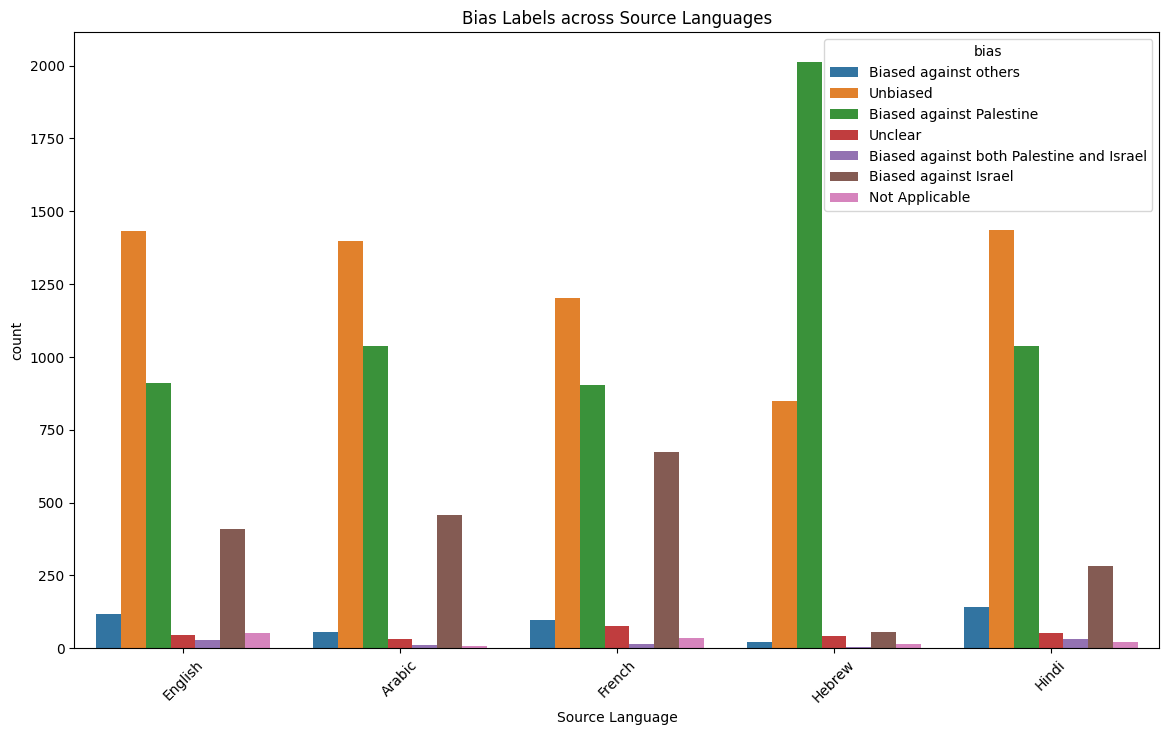}
\caption{Distribution of political bias categories in the human-annotated FIGNEWS subset.}
\label{fig:label_distribution}
\end{figure}

\subsection{Preprocessing}
We conducted extensive data preprocessing to ensure data consistency. All text was lowercased and stripped of URLs, emojis, and non-ASCII characters. Unlike traditional NLP pipelines, we retained stop words and inflections to preserve contextual richness which helps transformer-based models capture subtle rhetorical cues. Texts were normalized and tokenized using the \texttt{Hugging Face} tokenizer corresponding to each model. 

The preprocessed dataset was then inspected through exploratory data analysis (EDA), including frequency plots and word clouds to find key thematic terms and demographic distributions in the dataset.

\subsection{Data Augmentation}
The dataset exhibited substantial class imbalance, with two dominant categories (\textit{Unbiased}, \textit{Biased Against Palestine}) overshadowing the rest. In order to mitigate this, we employed GPT-4o-mini via the OpenAI API to paraphrase minority-class examples. For each underrepresented label, 3--5 semantically equivalent variants were generated per instance ensuring that the original meaning and bias polarity were preserved. This increased minority-class coverage and improved model robustness during fine-tuning.

All augmentation outputs were manually verified to ensure semantic consistency and avoid label drift.

\section{Methodology}

\subsection{Overview}
We employ both recurrent and transformer-based neural architectures to detect and quantify political bias in news articles. 
Our approach first establishes a sequence-modeling baseline using a BiLSTM network and subsequently fine-tunes several transformer models like BERT, RoBERTa, and LLaMA-3.1 on the annotated dataset.
Finally, we extend the framework by comparing human annotations with transformer-generated labels to assess cross-annotator consistency and model interpretability.

\subsection{Model Architectures}

\subsubsection{Baseline: BiLSTM}
To capture sequential dependencies and linguistic context, we implemented a Bidirectional Long Short-Term Memory (BiLSTM) network as a baseline.
The model processes text in both forward and backward directions which helps it to preserve contextual information across variable-length sequences. 
Word embeddings were initialized using pre-trained \textit{GloVe} and \textit{FastText} representations. 
The network comprises two stacked BiLSTM layers (512 and 256 units) followed by a dense layer with a softmax activation for seven-way classification.
Dropout layers and early stopping were applied to prevent overfitting, and optimization was performed using the \textit{Adam} optimizer with a learning rate of 0.001.

\subsubsection{Transformer Models}
For context-aware language representation, we employed transformer-based architectures—BERT \cite{devlin2019bert} and RoBERTa \cite{liu2019roberta}—as the core models for bias classification.
Both models were fine-tuned for sequence classification using the Hugging Face \texttt{Transformers} library. 
Special tokens (\texttt{[CLS]}, \texttt{[SEP]}) were used to encode sentence boundaries, and outputs from the \texttt{[CLS]} token were passed through a classification head.
We additionally experimented with \textit{LLaMA-3.1} (8B variant), fine-tuned using parameter-efficient tuning (QLoRA) and 4-bit quantization to evaluate the scalability of large language models under resource-constrained settings.

\subsection{Fine-Tuning Procedure}
Each model was fine-tuned using the same preprocessed dataset described in Section~\ref{sec:dataset}.
For transformers, we used a batch size of 64 and trained for three epochs with early stopping based on validation loss. 
Optimization was performed with \textit{AdamW} and a weight decay of 0.01.
Learning rates were set to $2 \times 10^{-5}$ for transformers and $1 \times 10^{-3}$ for the BiLSTM.
All experiments were conducted on an NVIDIA RTX~4090 GPU.
Cross-entropy loss was used for multi-class classification, and model checkpoints were saved based on the highest macro F1 score on the validation set.

\subsection{Evaluation Setup}
Model performance was evaluated on the held-out test set using accuracy, precision, recall, and macro F1-score. 
We also performed a paired \textit{t}-test to assess whether the difference between BERT and RoBERTa performance was statistically significant ($p > 0.05$).
Confusion matrices were used to analyze class-level misclassifications and identify systematic prediction errors.

\subsection{Human vs.~Machine Annotation Framework}
To evaluate how closely transformer-based models align with human perception of political bias, 
we designed an annotation framework comparing human-labeled data with automatically generated annotations
from both open-source and proprietary large language models (LLMs). 
This framework explores three inference regimes—\textbf{zero-shot}, \textbf{zero-shot with guideline conditioning}, 
and \textbf{few-shot with guideline examples}—across multiple model architectures.

\paragraph{Annotation Models}
We employed five representative language models covering a range of architectures and parameter scales:
\begin{itemize}  [itemsep=0pt, parsep=0pt, topsep=2pt, partopsep=0pt]
    \item \textbf{BERT-MNLI (base)}  a bidirectional transformer fine-tuned on the MultiNLI entailment task, 
    used as a lightweight baseline for entailment-based classification.
    \item \textbf{RoBERTa-MNLI (large)}  a robust transformer trained with improved optimization and dynamic masking, 
    serving as the primary zero-/few-shot classifier.
    \item \textbf{XLM-RoBERTa (large)}  a multilingual transformer enabling cross-lingual evaluation of the FIGNEWS dataset.
    \item \textbf{FLAN-T5 (base)}  a sequence-to-sequence model fine-tuned with instruction-following objectives.
    \item \textbf{ChatGPT~5.0 (API)}  a multimodal large language model with improved reasoning and context understanding
\end{itemize}

\paragraph{Annotation Regimes}
Each model was prompted under three distinct configurations:
\begin{enumerate}  [itemsep=0pt, parsep=0pt, topsep=2pt, partopsep=0pt]
    \item \textbf{Zero-shot:} Models classified texts into one of the seven bias categories without any external guidance. 
    \item \textbf{Zero-shot + Guidelines:} Models received concise natural language definitions for each bias label 
    (e.g., ``Biased against Palestine'' or ``Unbiased''), adapted from our human annotation manual.
    \item \textbf{Few-shot + Guidelines:} Models were provided with 2--3 human-annotated examples per label, 
    alongside guidelines serving as contextual anchors.
\end{enumerate}

\paragraph{Inference Pipeline.}
For open-source models, inference was conducted via the Hugging Face \texttt{transformers}\texttt{zero-shot-classification} pipeline,
using the \texttt{roberta-large-mnli} and \texttt{bert-base-mnli} checkpoints.
Each text was evaluated against all seven candidate hypotheses derived from the guideline definitions. 
The model produced entailment scores for each label, and the highest-scoring label was selected as the predicted bias category. 
To mitigate spurious predictions, additional decision rules were applied:
low-confidence predictions ($p < 0.22$) or minimal label margins ($\Delta p < 0.03$) were assigned as \textit{Unclear}, 
and texts deemed irrelevant to the Israel–Palestine conflict (via a secondary NLI-based relevance classifier) were labeled as \textit{Not Applicable}.  
A dual-threshold heuristic ($s_{BAI}, s_{BAP} \geq 0.30$, $|s_{BAI} - s_{BAP}| \leq 0.06$) was used to identify 
instances ``Biased against both Palestine and Israel.''

\paragraph{Guideline Integration.}
In guideline-conditioned setups, each hypothesis included the label’s definition, 
and for few-shot settings, short in-context examples drawn from the human-annotated training data. 
This combination improved interpretability and encouraged models to emulate human reasoning during bias labeling.
All prompts followed a unified structure:

\begin{quote}
\textit{``You are an expert annotator specializing in bias detection for social media news content about the Israel-Palestine conflict. Classify the [TEXT] into one of seven categories using these detailed guidelines [GUIDELINES]''}
\end{quote}

\paragraph{Outputs and Agreement Evaluation.}
Each model produced a machine-annotated dataset with predicted labels and confidence probabilities for each class.
These were aligned with human annotations to compute agreement metrics including Cohen’s $\kappa$, accuracy, precision, recall, and F1-score. 
Further qualitative evaluation examined systematic divergences between human and model decisions 
(e.g., sarcasm, implicit bias, or culturally dependent framing). 
This comparative framework enables a comprehensive assessment of how annotation consistency varies by model family, 
prompting regime, and the degree of human conditioning.

\section{Results}

\subsection{Performance of Fine-Tuned Models}

Table~\ref{tab:model-results} summarizes the classification performance of all fine-tuned models 
on the seven-way bias detection task. The transformer-based architectures substantially outperform 
the recurrent baseline, underscoring the benefit of contextual embeddings for political bias detection.

\begin{table}[h!]
\centering
\small
\setlength{\tabcolsep}{4pt}
\renewcommand{\arraystretch}{1.1}
\begin{tabular}{lccccc}
\hline
\textbf{Model} & \textbf{Accuracy} & \textbf{Precision} & \textbf{Recall} & \textbf{F1} & \textbf{Loss} \\
\hline
BiLSTM & 0.567 & 0.565 & 0.567 & 0.562 & 1.36 \\
BERT & 0.762 & 0.772 & 0.762 & 0.764 & 0.84 \\
RoBERTa & \textbf{0.772} & \textbf{0.785} & \textbf{0.772} & \textbf{0.774} & 0.81 \\
LLaMA-3.1 & 0.693 & 0.691 & 0.693 & 0.691 & \textbf{0.79} \\
\hline
\end{tabular}
\caption{Performance of fine-tuned models on the FIGNEWS dataset.
Metrics are reported on the held-out test set.
RoBERTa achieves the best overall performance across all metrics, 
while BiLSTM serves as a non-contextual baseline.}
\label{tab:model-results}
\end{table}

Across all models, RoBERTa achieved the highest macro F1-score (0.774), narrowly surpassing BERT (0.764).
The improvements in precision and recall demonstrate the effectiveness of dynamic masking 
and larger pretraining data in RoBERTa. 
While LLaMA-3.1 achieved competitive results given its quantized and parameter-efficient setup, 
it underperformed transformer baselines due to limited fine-tuning epochs and smaller batch sizes due to lack of computational resources. 
The BiLSTM model lagged behind, highlighting the importance of contextual representation learning 
for capturing subtle ideological cues.

A paired \textit{t}-test between BERT and RoBERTa outputs revealed that the performance difference 
was not statistically significant ($p > 0.05$), confirming both models’ strong and comparable reliability. 
However, performance differences across classes suggest that models identify overt polarity, but struggle with implicit or multi-actor framing, as shown by confusion concentrated among \textit{Unbiased}, \textit{Biased Against Palestine}, and \textit{Biased Against Israel} labels.
This indicates that the task is not purely lexical sentiment classification, but requires contextual interpretation that current models partially approximate.

\subsection{Human vs.~Machine Annotation Results}
Unlike the fine-tuned classifiers, zero-shot models, especially GPT-based, demonstrated moderate alignment with human judgments ($\kappa$ = 0.338), improving substantially when label definitions were included (guideline-conditioned prompting).
However, Zero-shot RoBERTa and BERT tended to over-predict “Unbiased”, suggesting a default neutrality stance when uncertainty is high. While FLAN-T5 showed higher sensitivity but low specificity, often detecting bias where humans did not.
This demonstrates that model behavior varies more by inference strategy (zero-shot vs. guided) than by model size alone.
\paragraph{Quantitative Agreement.}
Table~\ref{tab:human-machine-wide} presents agreement metrics across models and prompting regimes.
Among the non-generative models, \textbf{RoBERTa-MNLI (zero-shot + guideline)} achieved the strongest alignment with human labels (Accuracy $=0.419$, $\kappa=0.133$), along with the lowest Jensen–Shannon divergence, indicating that guideline conditioning not only improves classification agreement but also calibrates the predicted label distribution toward human-like judgments.
This  steady improvement across architectures shows that \textit{explicit operationalization of bias categories reduces ambiguity and promotes more stable decision boundaries}.

\begin{table*}[t]
\centering
\small
\setlength{\tabcolsep}{6pt}
\renewcommand{\arraystretch}{1.08}
\begin{tabular}{l l c c c c}
\toprule
Model & Setup & Accuracy & Matthews corrcoef & $\kappa$ & Jensen–Shannon divergence\\
\midrule
BERT-MNLI & Zero-shot & 0.094 & 0.068 & 0.015 & 0.601 \\
BERT-MNLI & Zero-shot +Guideline & 0.155 & 0.101 & 0.033 & 0.496 \\
BERT-MNLI & Few-shot +Guideline & 0.188 & 0.090 & 0.039 & 0.524 \\
RoBERTa-MNLI & Zero-shot & 0.128 & 0.105 & 0.049 & 0.581 \\
RoBERTa-MNLI & Zero-shot +Guideline & \textbf{0.419} & \textbf{0.198} & \textbf{0.133} & \textbf{0.179} \\
RoBERTa-MNLI & Few-shot +Guideline & 0.062 & 0.067 & 0.009 & 0.656 \\
FLAN-T5 & Zero-shot +Guideline & 0.296 & 0.179 & 0.055 & 0.294 \\
FLAN-T5 & Few-shot +Guideline & 0.322 & 0.177 & 0.055 & 0.233 \\
XLM-R (large) & Zero-shot +Guideline & 0.097 & 0.069 & 0.013 & 0.614 \\
GPT-5.0 & Zero-shot & \textbf{0.553} & \textbf{0.350} & \textbf{0.338} & \textbf{0.137} \\
\bottomrule
\end{tabular}
\caption{Agreement between human and model annotations. 
Metrics computed on the shared training set ($N{=}2961$). 
JS divergence measures label distribution distance (lower is better).}
\label{tab:human-machine-wide}
\end{table*}

\paragraph{Label-wise Trends.}
All models had highest agreement with human annotators for \textit{Unbiased} and \textit{Biased against Palestine} labels and consistently struggled with \textit{Biased against Both} and \textit{Unclear} labels. This reflects the broader challenge of detecting implicit stance and multi-actor framing that require contextual reasoning beyond lexical or entity sentiment cues \cite{spinde2021babe}.

\paragraph{Interpretation.}
Collectively, the results show that \textit{accuracy alone obscures significant differences in annotation reasoning}. Guideline-conditioned prompting encourages models to adopt more human-like labels that reduce both over-neutral and over-biased predictions. However, the decline in performance in the few-shot setting especially for RoBERTa suggests that non-instruction-tuned models may treat in-context examples as noise when they are not aligned with their pretraining objectives. Meanwhile, FLAN-T5's degradation under few-shot prompting indicates a tendency toward surface-level pattern matching rather than conceptual generalization. These findings reinforce that human–machine alignment is shaped not only by model capacity but by how task semantics are represented and communicated during inference.

\section{Result Analysis}

\subsection{Overview}
The results reveal systematic differences in how models interpret political bias. Fine-tuned transformer architectures, particularly RoBERTa, achieved higher supervised performance than BiLSTM and zero-shot baselines. However, improvements in accuracy and macro-F1 were primarily driven by recognition of \textit{overt sentiment polarity} rather than discourse-level stance reasoning. Even the strongest models struggled with implicit, multi-actor, and context-dependent bias, indicating that current approaches approximate bias via lexical cues rather than interpretive inference.

\subsection{Analysis of Fine-Tuned RoBERTa Behavior}
RoBERTa consistently outperformed BERT and BiLSTM in accuracy (0.772 vs.\ 0.762 and 0.567, respectively) and macro-F1 (0.774 vs.\ 0.764 and 0.562), confirming the benefits of contextualized embeddings and dynamic masking. Nevertheless, qualitative inspection revealed several recurring weaknesses.

\paragraph{Lexical and Framing Sensitivity}
RoBERTa frequently associated emotionally charged terms (e.g., \textit{“terrorist”}, \textit{“occupation”}) with directional bias regardless of narrative framing. Statements such as \textit{``There are terrorists in my house''} were misclassified as \textit{Biased against Palestine}, demonstrating reliance on lexical polarity rather than contextual attribution.

\paragraph{Entity-Centric Bias Inference}
Bias direction was often inferred from which actor appeared in negative proximity. Mentions of Israeli state actions skewed predictions toward \textit{Biased against Palestine}, while references to Hamas skewed toward \textit{Biased against Israel}. Balanced statements with explicit dual stances were frequently collapsed into singular polarity.

\paragraph{Implicit and Contextual Bias}
RoBERTa failed to infer bias expressed indirectly through questioning, critique, implication, or omission. For example, \textit{``Gaza, a ground offensive doomed to failure?''} was labeled as \textit{Biased against Israel} despite being an internal policy critique rather than ideological stance.

\paragraph{Ambiguity and Attribution}
Quoted speech was often treated as authorial opinion. Headlines attributing statements to political actors (\textit{``Netanyahu: Israel will return to fighting...''}) were labeled as biased, indicating absence of speaker–narrator distinction. The model also defaulted to “Biased against both” under uncertainty, suggesting conservative over-generalization. These errors point to the need for discourse-aware objectives and attribution-sensitive fine-tuning.

\paragraph{Summary}
RoBERTa’s fine-tuning improved explicit bias detection but exposed enduring weaknesses in neutrality recognition and mixed-stance comprehension. Its improvements reflect stronger lexical precision and not genuine stance reasoning. The model fails to interpret multi-source narratives, attribution, or  how framing works, which are key parts of political bias.

\subsection{Human–Machine Divergence (RoBERTa-MNLI)}
Comparison between human and RoBERTa-MNLI (zero-shot with guidelines) annotations show the persistent gap between rule-conditioned models and human inference. The model achieved 41.9\% accuracy and $kappa$  = 0.13 which is only slight agreement beyond chance. It under-detected bias in over half of “Biased against Palestine” cases and 42\% of “Biased against Israel” cases, often labeling them “Unbiased.” Conversely, it over-predicted bias in one-third of truly neutral examples that highlight asymmetry between sensitivity and specificity.  
Distributional analysis showed overuse of “Unbiased” (57\% vs.\ 41\% in human labels) and underuse of “Unclear” or “Not Applicable,” with a Jensen–Shannon divergence of 0.046. This imbalance reflects over-confidence in assigning directional bias even under semantic uncertainty.

\paragraph{Qualitative Divergence}
Representative examples reveal systematic misalignment:
(1) factual event descriptions misinterpreted as biased due to lexical triggers;
(2) overtly partisan hashtags dismissed as “Unclear”; and
(3) off-topic statements forced into directional labels.  
These errors reinforce that zero-shot transformers rely on surface cues instead of evaluative intent which confirms findings from prior alignment work \cite{Plank2022,movva2024alignment}.

\subsection{GPT-5 Zero-Shot Performance}
The GPT-5 model achieved 55.3\% accuracy ($kappa $ = 0.34, weighted-F1 = 0.57), outperforming RoBERTa-MNLI and XLM-R on the same task. Without explicit supervision, GPT-5 effectively captured broad sentiment polarity and overt bias cues, particularly for high-frequency categories such as “Unbiased” (F1 = 0.62) and “Biased against Palestine” (F1 = 0.62).  
However, the model exhibited a critical tendency to default to the "Unbiased" classification when presented with nuanced or low-resource classes (like "Biased against Others" or "Unclear"). This preference for neutrality under conditions of uncertainty, rather than making a categorical commitment, reflects the model's underlying mechanism of probabilistic averaging, a behavior likely rooted in its instruction-tuning safety optimizations.

\paragraph{Cross-Model Comparison}
GPT-5 was better at maintaining contextual consistency when dealing with instances that involved multiple actors than RoBERTa, which tended to rely too much on basic rules for identifying entities. However, GPT-5's overall interpretation of the text varied more widely. While RoBERTa performed well due to its specialized training for precise word choice, GPT-5's main advantage was its ability to perform tasks it hadn't been explicitly trained on (zero-shot adaptability) and its broad applicability across different subjects. Nevertheless, GPT-5's lack of specific training for particular tasks meant it struggled to be consistent when handling unclear or ambiguous situations.

\subsection{GPT-5 Variant Analysis}
\begin{table}[h!]
\centering
\footnotesize
\renewcommand{\arraystretch}{1.2} 
\begin{tabular}{lccccc}
\hline
\textbf{Model} & \textbf{Acc.} & \textbf{M-F1} & \textbf{W-F1} & \textbf{$\kappa$}  & \textbf{Sup.} \\
\hline
GPT-5 (ZS) & 0.68 & 0.12 & 0.77 & 0.09 &  877 \\
GPT-5 (ZS+GL) & 0.63 & 0.13 & 0.72 & 0.08  & 656 \\
GPT-5 (FS+GL) & 0.58 & 0.13 & 0.69 & 0.07 & 714 \\
\hline
\end{tabular}
\caption{Performance comparison of GPT-5 variants on a sample subset. 
\textbf{ZS} = Zero-Shot, \textbf{GL} = Guideline, \textbf{FS} = Few-Shot.}
\end{table}

Only a subset of 600–700 samples was tested across GPT-5 configurations. It is important to note that the evaluated subset was highly imbalanced, with the majority of samples labeled as \textit{Unbiased} or \textit{Unclear}, while the explicitly biased categories were sparsely represented. Zero-shot GPT outperformed guideline- and few-shot variants, suggesting that additional instruction constrained its broader representational inference and introduced label conservatism. This reflects a known effect of safety-aligned instruction tuning, where models are discouraged from strong interpretive claims unless high certainty is present.

\subsection{Where Humans and Models Diverge}

Across all models and inference settings, three systematic divergence patterns emerged between human and machine bias perception (Table~\ref{tab:divergence-table}). These patterns indicate that even when label agreement appears moderate, the \textit{underlying reasoning strategies} differ substantially.

\begin{table}[h!]
\centering
\footnotesize
\renewcommand{\arraystretch}{1.15}
\begin{tabular}{p{2.2cm} p{2.2cm} p{2.2cm}}
\toprule
\textbf{Divergence Type} & \textbf{Model Behavior} & \textbf{Human Interpretation} \\
\midrule
Lexical sentiment cues &
Bias inferred from emotionally loaded words (e.g., \textit{terrorist''}, \textit{occupation''}) regardless of context &
Bias judged based on \textit{source attribution, intent}, and \textit{narrative framing}, not single-word polarity \\
\midrule
Entity association heuristics &
Bias direction often tied to which actor receives negative descriptive language &
Humans evaluate speaker identity, quoted sources, and rhetorical stance before inferring bias \\
\midrule
Implicit / multi-actor framing &
Ambiguous or balanced texts frequently labeled as \textit{Unbiased} &
Humans infer stance through emphasis, omission, and relational framing across actors \\
\bottomrule
\end{tabular}
\caption{Systematic reasoning differences between human and model bias judgments.}
\label{tab:divergence-table}
\end{table}

These divergences show that current models primarily rely on surface-level lexical and entity-association cues, whereas human annotators use contextual, pragmatic, and discourse-level reasoning.

Thus, even when overall accuracy is high, models fail to detect genuine political bias. \textbf{Ultimately, models are not detecting political bias itself, they are detecting local sentiment polarity toward named entities.}

This methodological discrepancy explains several systematic failure modes observed in model performance:
\begin{itemize}[itemsep=0pt, parsep=0pt, topsep=2pt, partopsep=0pt]
\item misclassification of neutral but emotionally toned statements,

\item failure on multi-actor narratives, and

\item over-reliance on words rather than narrative framing.
\end{itemize}
This analysis strongly motivates the need for a new generation of models capable of \textit{representing political stance as a comprehensive discourse structure}, moving decisively beyond mere token-level sentiment analysis.

\subsection{Comparative Insights}
Synthesizing across models, several trends emerge:
fine-tuned transformer models offer stability but exhibit only surface-oriented performance, while Large Language Models (LLMs) are more adaptable yet prone to interpretive drift. The core differences between human and machine analysis typically revolve around the models' inability to handle subtle contextualization, accurately manage quotation handling, and detect implicit bias. Ultimately, increasing model scale only improves the coverage or breadth of detection, not the interpretive fidelity. Closing this critical gap requires developing hybrid frameworks that integrate human-level contextual reasoning, advanced discourse-level modeling, and specific, bias-aware objectives to ensure a robust and trustworthy system for political-bias detection.

\section{Limitations}
While the present study offers meaningful insights into human–model divergence in bias perception, several limitations should be noted. First, the research is focused exclusively on the Israel-Palestine conflict, which is highly complex and emotional, meaning the results might not apply to other political topics. Second, since political bias is subjective, the human-assigned labels used represent a consensus among annotators, not an absolute, single truth. Third, computational constraints restricted fine-tuning experiments to medium-scale models; larger architectures may exhibit different alignment behavior. Finally, the study only examined textual data, neglecting how bias is often amplified or conveyed through multimodal elements like images and overall network effects. Addressing these factors presents a valuable direction for future research.

\section{Future Work}
Future work will expand this framework beyond a single conflict domain to more diverse datasets and languages, testing cross-cultural robustness. With greater computational resources, larger LLMs and ensemble approaches can be evaluated to examine the effect of model scale on human–machine alignment. We also plan to incorporate discourse-aware objectives and entity-relation modeling to capture implicit and multi-actor bias. Integrating explainability and human-in-the-loop calibration could further improve interpretability. Finally, multimodal extensions combining textual, visual, and social-network signals may offer a richer understanding of media bias dynamics.

\section{Conclusion}
This study provides a comparative analysis of political bias perception across human annotators and multiple language model architectures. While RoBERTa achieved the highest supervised performance and GPT-based models demonstrated strong zero-shot adaptability, our analysis shows that alignment in labels does not equate to alignment in reasoning. Transformer models primarily detect lexical and sentiment cues, whereas human annotators interpret bias through contextual framing, attribution, and implied stance. These divergences suggest that improving automated bias detection requires moving beyond surface-level text patterns toward discourse-level reasoning and narrative structure modeling. Future work should focus on incorporating multi-actor relational context, rhetorical framing cues, and interpretability-driven training objectives to bridge the gap between human and model bias interpretation.

\section{Ethics Statement}
All data were obtained from the publicly available FIGNEWS 2024 corpus under its research license. No personally identifiable information was used. Annotators were briefed on ethical considerations and potential emotional impact due to conflict-related content.



\bibliography{custom}

@inproceedings{baly2020detect,
    title = "We Can Detect Your Bias: Predicting the Political Ideology of News Articles",
    author = "Baly, Ramy  and
      Da San Martino, Giovanni  and
      Glass, James  and
      Nakov, Preslav",
    editor = "Webber, Bonnie  and
      Cohn, Trevor  and
      He, Yulan  and
      Liu, Yang",
    booktitle = "Proceedings of the 2020 Conference on Empirical Methods in Natural Language Processing (EMNLP)",
    month = nov,
    year = "2020",
    address = "Online",
    publisher = "Association for Computational Linguistics",
    url = "https://aclanthology.org/2020.emnlp-main.404/",
    doi = "10.18653/v1/2020.emnlp-main.404",
    pages = "4982--4991"
}

@inproceedings{fan2019hierarchical,
  title={Hierarchical Neural Network for Stance and Bias Detection},
  author={Fan, Fan and Zhou, Jiaqi and Wu, Yifan and Li, Chenliang},
  booktitle={Proceedings of the 2019 Conference on Empirical Methods in Natural Language Processing},
  pages={3405--3414},
  year={2019},
  publisher={Association for Computational Linguistics},
  doi={10.18653/v1/D19-1340}
}

@misc{morstatter2018bias,
      title={When is it Biased? Assessing the Representativeness of Twitter's Streaming API}, 
      author={Fred Morstatter and Jürgen Pfeffer and Huan Liu},
      year={2014},
      eprint={1401.7909},
      archivePrefix={arXiv},
      primaryClass={cs.SI},
      url={https://arxiv.org/abs/1401.7909}, 
}

@misc{feng2023pretrainingdatalanguagemodels,
      title={From Pretraining Data to Language Models to Downstream Tasks: Tracking the Trails of Political Biases Leading to Unfair NLP Models}, 
      author={Shangbin Feng and Chan Young Park and Yuhan Liu and Yulia Tsvetkov},
      year={2023},
      eprint={2305.08283},
      archivePrefix={arXiv},
      primaryClass={cs.CL},
      url={https://arxiv.org/abs/2305.08283}, 
}

@inproceedings{jaremko2025implicit,
    title = "Revisiting Implicitly Abusive Language Detection: Evaluating {LLM}s in Zero-Shot and Few-Shot Settings",
    author = "Jaremko, Julia  and
      Gromann, Dagmar  and
      Wiegand, Michael",
    editor = "Rambow, Owen  and
      Wanner, Leo  and
      Apidianaki, Marianna  and
      Al-Khalifa, Hend  and
      Eugenio, Barbara Di  and
      Schockaert, Steven",
    booktitle = "Proceedings of the 31st International Conference on Computational Linguistics",
    month = jan,
    year = "2025",
    address = "Abu Dhabi, UAE",
    publisher = "Association for Computational Linguistics",
    url = "https://aclanthology.org/2025.coling-main.262/",
    pages = "3879--3898"
}

@inproceedings{movva2024alignment,
  title={Annotation Alignment: Comparing LLM and Human Annotations of Conversational Safety},
  author={Movva, Rajiv and Koh, Pang Wei and Pierson, Emma},
  booktitle={Proceedings of the 2024 Conference on Empirical Methods in Natural Language Processing},
  pages={9048--9062},
  year={2024},
  publisher={Association for Computational Linguistics},
  doi={10.18653/v1/2024.emnlp-main.511}
}

@inproceedings{muenker2025zeroshot,
      title={Zero-shot prompt-based classification: topic labeling in times of foundation models in German Tweets}, 
      author={Simon Münker and Kai Kugler and Achim Rettinger},
      year={2024},
      eprint={2406.18239},
      archivePrefix={arXiv},
      primaryClass={cs.CL},
      url={https://arxiv.org/abs/2406.18239}, 
}

@inproceedings{chiang2023evaluation,
  title={Can Large Language Models Be an Alternative to Human Evaluations?},
  author={Chiang, Cheng-Han and Lee, Hung-yi},
  booktitle={Proceedings of the 61st Annual Meeting of the Association for Computational Linguistics},
  pages={15607--15631},
  year={2023},
  publisher={Association for Computational Linguistics},
  doi={10.18653/v1/2023.acl-long.870}
}

@inproceedings{gilardi2023gpt,
   title={ChatGPT outperforms crowd workers for text-annotation tasks},
   volume={120},
   ISSN={1091-6490},
   url={http://dx.doi.org/10.1073/pnas.2305016120},
   DOI={10.1073/pnas.2305016120},
   number={30},
   journal={Proceedings of the National Academy of Sciences},
   publisher={Proceedings of the National Academy of Sciences},
   author={Gilardi, Fabrizio and Alizadeh, Meysam and Kubli, Maël},
   year={2023},
   month=jul }

@article{davani2022disagreement,
  title={Dealing with Disagreements: Looking Beyond the Majority Vote in Subjective Annotations},
  author={Davani, Aida Mostafazadeh and Diaz, Mark and Prabhakaran, Vinodkumar},
  journal={Transactions of the Association for Computational Linguistics},
  volume={10},
  pages={92--110},
  year={2022},
  doi={10.1162/tacl_a_00449}
}

@inproceedings{plank2022human,
  title={The “Problem” of Human Label Variation: On Ground Truth in Data, Modeling and Evaluation},
  author={Plank, Barbara},
  booktitle={Proceedings of the 2022 Conference on Empirical Methods in Natural Language Processing},
  pages={10671--10682},
  year={2022},
  publisher={Association for Computational Linguistics},
  doi={10.18653/v1/2022.emnlp-main.731}
}

@misc{santurkar2023opinions,
      title={Whose Opinions Do Language Models Reflect?}, 
      author={Shibani Santurkar and Esin Durmus and Faisal Ladhak and Cinoo Lee and Percy Liang and Tatsunori Hashimoto},
      year={2023},
      eprint={2303.17548},
      archivePrefix={arXiv},
      primaryClass={cs.CL},
      url={https://arxiv.org/abs/2303.17548}, 
}

@inproceedings{nie2024multilingual,
    title = "Do Multilingual Large Language Models Mitigate Stereotype Bias?",
    author = {Nie, Shangrui  and
      Fromm, Michael  and
      Welch, Charles  and
      G{\"o}rge, Rebekka  and
      Karimi, Akbar  and
      Plepi, Joan  and
      Mowmita, Nazia  and
      Flores-Herr, Nicolas  and
      Ali, Mehdi  and
      Flek, Lucie},
    editor = "Prabhakaran, Vinodkumar  and
      Dev, Sunipa  and
      Benotti, Luciana  and
      Hershcovich, Daniel  and
      Cabello, Laura  and
      Cao, Yong  and
      Adebara, Ife  and
      Zhou, Li",
    booktitle = "Proceedings of the 2nd Workshop on Cross-Cultural Considerations in NLP",
    month = aug,
    year = "2024",
    address = "Bangkok, Thailand",
    publisher = "Association for Computational Linguistics",
    url = "https://aclanthology.org/2024.c3nlp-1.6/",
    doi = "10.18653/v1/2024.c3nlp-1.6",
    pages = "65--83"
}

@article{devlin2019bert,
  title     = {BERT: Pre-training of Deep Bidirectional Transformers for Language Understanding},
  author    = {Devlin, Jacob and Chang, Ming-Wei and Lee, Kenton and Toutanova, Kristina},
  journal   = {arXiv preprint arXiv:1810.04805},
  year      = {2019},
  url       = {https://arxiv.org/abs/1810.04805}
}

@article{liu2019roberta,
  title     = {RoBERTa: A Robustly Optimized BERT Pretraining Approach},
  author    = {Liu, Yinhan and Ott, Myle and Goyal, Naman and Du, Jingfei and Joshi, Mandar and Chen, Danqi and Levy, Omer and Lewis, Mike and Zettlemoyer, Luke and Stoyanov, Veselin},
  journal   = {arXiv preprint arXiv:1907.11692},
  year      = {2019},
  url       = {https://arxiv.org/abs/1907.11692}
}

@article{zaghouani2024fignews,
  title     = {The FIGNEWS Shared Task on News Media Narratives},
  author    = {Zaghouani, Wajdi and Jarrar, Mustafa and Habash, Nizar and et al.},
  journal   = {arXiv preprint arXiv:2407.18147},
  year      = {2024},
  url       = {https://arxiv.org/abs/2407.18147}
}

@inproceedings{spinde2021babe,
    title = "Neural Media Bias Detection Using Distant Supervision With {BABE} - Bias Annotations By Experts",
    author = "Spinde, Timo  and
      Plank, Manuel  and
      Krieger, Jan-David  and
      Ruas, Terry  and
      Gipp, Bela  and
      Aizawa, Akiko",
    editor = "Moens, Marie-Francine  and
      Huang, Xuanjing  and
      Specia, Lucia  and
      Yih, Scott Wen-tau",
    booktitle = "Findings of the Association for Computational Linguistics: EMNLP 2021",
    month = nov,
    year = "2021",
    address = "Punta Cana, Dominican Republic",
    publisher = "Association for Computational Linguistics",
    url = "https://aclanthology.org/2021.findings-emnlp.101/",
    doi = "10.18653/v1/2021.findings-emnlp.101",
    pages = "1166--1177"
}

@misc{fignews2024,
  title        = {FIGNEWS 2024: News Media Narratives Dataset},
  author       = {{FIGNEWS}},
  year         = {2024},
  howpublished = {\url{https://sites.google.com/view/fignews/home}},
  note         = {Part of the The Second Arabic Natural Language Processing Conference
(ArabicNLP 2024) Co-located with ACL 2024}
}

@inproceedings{plank2022,
    title = "The ``Problem'' of Human Label Variation: On Ground Truth in Data, Modeling and Evaluation",
    author = "Plank, Barbara",
    editor = "Goldberg, Yoav  and
      Kozareva, Zornitsa  and
      Zhang, Yue",
    booktitle = "Proceedings of the 2022 Conference on Empirical Methods in Natural Language Processing",
    month = dec,
    year = "2022",
    address = "Abu Dhabi, United Arab Emirates",
    publisher = "Association for Computational Linguistics",
    url = "https://aclanthology.org/2022.emnlp-main.731/",
    doi = "10.18653/v1/2022.emnlp-main.731",
    pages = "10671--10682"
}

@article{rodrigo2024media,
title = {A systematic review on media bias detection: What is media bias, how it is expressed, and how to detect it},
journal = {Expert Systems with Applications},
volume = {237},
pages = {121641},
year = {2024},
issn = {0957-4174},
doi = {https://doi.org/10.1016/j.eswa.2023.121641},
url = {https://www.sciencedirect.com/science/article/pii/S0957417423021437},
author = {Francisco-Javier Rodrigo-Ginés and Jorge Carrillo-de-Albornoz and Laura Plaza}
}

@article{lee2023opiniongaps,
  title={Mind the Opinion Gap: Understanding What LLMs Miss in Capturing Human Disagreement},
  author={Lee, Cinoo and Santurkar, Shibani and Ladhak, Faisal and Hashimoto, Tatsunori},
  journal={arXiv preprint arXiv:2310.01980},
  year={2023},
  url={https://arxiv.org/abs/2310.01980}
}

@misc{bojic2025,
      title={Evaluating Large Language Models Against Human Annotators in Latent Content Analysis: Sentiment, Political Leaning, Emotional Intensity, and Sarcasm}, 
      author={Ljubisa Bojic and Olga Zagovora and Asta Zelenkauskaite and Vuk Vukovic and Milan Cabarkapa and Selma Veseljević Jerkovic and Ana Jovančevic},
      year={2025},
      eprint={2501.02532},
      archivePrefix={arXiv},
      primaryClass={cs.CL},
      url={https://arxiv.org/abs/2501.02532}, 
}

\appendix

\section{Appendix}
\subsection{Guidelines}
\label{sec:appendix}

\textbf{Unbiased:} Content labeled “Unbiased” presents information objectively, avoiding favoritism or prejudice. While some word choices might hint at the author’s per spective, the content remains unbiased if these choices are contextually relevant and don’t significantly distort the overall narrative. An article is considered unbiased if it offers fair representation of all sides, avoids misleading language, and doesn’t express opinions favoring a particular viewpoint. For example, the statement “Both parties have expressed their willingness to nego tiate, according to international mediators” presents the viewpoints of both sides equally, without favoring one over the other. Attributing the information to an ex 21 ternal source (international mediators) further enhances neutrality. Similarly, the sentence “The Palestinian death toll in Gaza from the war between Israel and the territory’s Hamas rulers has soared past 25,000, the Gaza Health Ministry said” reports the death toll neutrally, avoiding blame or bias. Attributing the information to the Hamas-controlled health ministry ensures that subjective content is sourced externally. The statement “South Africa has filed an application at the International Court of Justice to begin proceedings over allegations of genocide against Israel” is also neutral, as it attributes the information to the International Court of Justice. The facts are presented without bias, allowing the audience to form their own conclusions. 
 
 \textbf{Unclear:} The label “Unclear” is applied to those articles where it is difficult to determine the bias due to vague language, incomplete information or insufficient context. These articles often discuss sensitive issues without clearly identifying which side of the argument they are supporting or opposing. This can happen in media coverage of conflicts where the language avoids direct attribution on purpose. For instance, the quote “ The story of the battlefield... in the words of ZEE NEWS The great victory of ‘Operation Ajay’ Watch the time fixed to meet the soil of ‘Gaza’ \#Deshhit LIVE with \#ChandanSingh \#IsraelHamasWar \#Gaza \#IsraelArmy \#OperationAjay
 \#ZeeNews \#ZeeLive” mentions a battlefield and a success ful operation, but doesn’t say who was involved, or the broader context. It’s difficult to determine whether the statement carries bias or not or who it might be biased against. Additionally, texts or titles that are too short, such as “Losing support” or “He was a real hero” are labeled as UNCLEAR. These texts may have links to other media or contain too little information to evaluate properly. 
 
 \textbf{Biased against Palestine:} A text was labeled biased against Palestine if the text unfairly omitted or downplayed Palestinian viewpoints, had only one sided rep resentation, used emotionally charged or pejorative language against Palestinians such as referring them as terrorists, exhibited facts that support a negative image of Palestine while ignoring justification and legitimate grievances, focused on negative stereotypes and generalizations of the Palestinians. For example: “LIVE: Presi dent Joe Biden Responds to Hamas Terrorist Attack on Israel…”. Here the label “terrorist” without providing context for Palestinian grievances reflects a one-sided portrayal, leading to bias. Another instance: “The singer Idan Raichel attacked: “The Palestinians are not rebelling against Hamas- therefore most of them should be treated as involved.”” “They could have been brave, entered the tunnels tonight and revolted,” added Reichel”. In this text all Palestinians have been generalized as complicit for not rebelling against Hamas which unfairly stereotypes and dehuman izes the broader population.
 
 \textbf{Biased against Israel:} This label has been used for texts which portrayed Israel in a negative or distorted manner such as Zionist or occupation, used emotionally charged language to describe Israel without context or explanation, highlighted Is raeli military actions without justification, portrayed Israelis in a way that dehuman izes them or justifies violence against them, disregarded Israel’s legitimate concerns regarding security and unfairly portrayed Israeli actions as purely aggressive. Ex 22 ample: “Israeli gunboats violate the truce and bomb the coast of Khan Yunis”. This text focuses solely on Israeli military actions, portraying them as aggressive without context or explanation of potential security concerns. Another example: “Urgent | “Osama Hamdan, a leader in Hamas”:- “The Israeli aggression has written its mili tary and political end”- “The occupation leaders will not see their prisoners held by the resistance alive until after a comprehensive cessation of the aggression.””. This text refers to Israel’s actions as “aggression” and “occupation” without acknowledg ing Israel’s security context, dehumanizing Israeli leadership and justifying violence against them. 
 
 \textbf{Biased against both Palestine and Israel:} The label “Biased against both Palestine and Israel” is used for articles that paint both sides of the conflict in an unfavorable light, often ridiculing or undermining the actions of both Palestine and Israel. For example, the statement “LIVE | Sea, land, sky... Israel and Hamas clashed, Hamas fired 5 thousand rockets... Israel opened ammunition stockpile” suggests both parties are engaged in harmful actions, without favoring one side. Another example is “In the latest chapter of their endless and futile conflict, Israeli and Palestinian forces have once again engaged in senseless violence, showing a com plete disregard for peace or human life.” This type of article scrutinizes the actions of both sides and expresses opinions that both sides’ irresponsibility is the cause for the breakdown of peace. The media supports a narrative that suggests both parties are equally responsible for the violence and instability by portraying the dispute in this way. 
 
 \textbf{Biased against others:} A text has been labeled as “Biased against others” if it portrayed entities other than Israel or Palestine in a negative, distorted, or unfair manner by misrepresenting their roles or intentions in the Israel-Palestine conflict. For instance: “Iranian President Raisi meets Putin, accuses US of genocide in Gaza.” This text criticizes the US with loaded language like “genocide” without providing balanced or factual evidence, leading to misrepresentation of its role. Similarly the text “France, Germany, England, Austria, the entire EU, the USA... are responsible for the bombing of a hospital in Gaza by the Israeli army!!..” he text unfairly blames multiple international actors for an Israeli military action, misrepresenting their in volvement in the conflict and promoting bias against them. Such texts are labeled “Biased against others” when they target groups or countries apart from Israel and Palestine with unfair or emotionally charged portrayals. 
 
 \textbf{Not Applicable:} Articles or headlines have been labeled as “Not Applicable” if the text didn’t pertain to the Israel-Palestine conflict. These texts contained unrelated topics which didn’t require any bias analysis and even if the texts were biased, they were irrelevant to the Israel Palestine conflict or didn’t mention Israel, Palestine, or any entities involved in the conflict. For instance: “Karine Jean-Pierre briefs the press alongside John Kirby.”, “In the United Kingdom, the BBC is under fire: red paint was even sprayed on the facade of the group.”

\end{document}